\documentclass[preprint,review,12pt]{elsarticle}

\usepackage{amssymb}
\usepackage{graphicx}
\usepackage{amsmath}
\usepackage{amssymb}
\usepackage{booktabs}
\usepackage{float}
\usepackage{multirow}
\usepackage{url}
\usepackage[colorlinks=True,
            linkcolor=red,
            anchorcolor=blue,
            citecolor=green]{hyperref}
            
\usepackage{stfloats}
\usepackage{makecell, tabularx}
\usepackage{lineno}
\modulolinenumbers[5]

\journal{XXXX}

\begin{document}

\begin{frontmatter}

%% Title, authors and addresses

%% use the tnoteref command within \title for footnotes;
%% use the tnotetext command for theassociated footnote;
%% use the fnref command within \author or \address for footnotes;
%% use the fntext command for theassociated footnote;
%% use the corref command within \author for corresponding author footnotes;
%% use the cortext command for theassociated footnote;
%% use the ead command for the email address,
%% and the form \ead[url] for the home page:
%% \title{Title\tnoteref{label1}}
%% \tnotetext[label1]{}
%% \author{Name\corref{cor1}\fnref{label2}}
%% \ead{email address}
%% \ead[url]{home page}
%% \fntext[label2]{}
%% \cortext[cor1]{}
%% \affiliation{organization={},
%%             addressline={},
%%             city={},
%%             postcode={},
%%             state={},
%%             country={}}
%% \fntext[label3]{}

\title{LLaMA-Reg: Using LLaMA 2 for Unsupervised Medical Image Registration}

%% use optional labels to link authors explicitly to addresses:
%% \author[label1,label2]{}
%% \affiliation[label1]{organization={},
%%             addressline={},
%%             city={},
%%             postcode={},
%%             state={},
%%             country={}}
%%
%% \affiliation[label2]{organization={},
%%             addressline={},
%%             city={},
%%             postcode={},
%%             state={},
%%             country={}}

\author[label1]{Mingrui Ma\fnref{equal}}
\ead{mamr@kust.edu.cn}

\author[label2,label3,label4]{Yu Yang\fnref{equal}\corref{Corresponding}}
\ead{yangy6874@enzemed.com}
\cortext[Corresponding]{Corresponding author.}
\fntext[equal]{These authors contributed equally to this work.}

\affiliation[label1]{organization={Faculty of Information Science Engineering and Automation, Kunming University of Science and Technology},%Department and Organization
            city={Kunming},
            country={China}}

\affiliation[label2]{organization={Department of Orthopaedics, Taizhou Hospital of Zhejiang Province Affiliated with Wenzhou Medical University},%Department and Organization
city={Wenzhou},
country={China}}

\affiliation[label3]{organization={Department of Orthopaedics, Enze Hospital, Taizhou Enze Medical Centre (Group)},%Department and Organization
city={Taizhou},
country={China}}

\affiliation[label4]{organization={Department of Orthopaedics, The Second Affiliated Hospital of Dalian Medical University},%Department and Organization
city={Dalian},
country={China}}

\begin{abstract}
Medical image registration is an essential topic in medical image analysis. In this paper, we propose a method for medical image registration using a pretrained large language model. We find that using the pretrained large language model to encode deep features of the medical images in the registration model can effectively improve image registration accuracy, indicating the great potential of the large language model in medical image registration tasks. We use dual encoders to perform deep feature extraction on image pairs and then input the features into the pretrained large language model. To adapt the large language model to our registration task, the weights of the large language model are frozen in the registration model, and an adapter is utilized to fine-tune the large language model, which aims at (a) mapping the visual tokens to the language space before the large language model computing, (b) project the modeled language tokens output from the large language model to the visual space. Our method combines output features from the fine-tuned large language model with the features output from each encoder layer to gradually generate the deformation fields required for registration in the decoder. To demonstrate the effectiveness of the large prediction model in registration tasks, we conducted experiments on knee and brain MRI and achieved state-of-the-art results.

\end{abstract}

% %%Graphical abstract
% \begin{graphicalabstract}
% %\includegraphics{grabs}
% \end{graphicalabstract}

%%Research highlights
% \begin{highlights}
% \item Research highlight 1
% \item Research highlight 2
% \end{highlights}


\end{frontmatter}

% \linenumbers

%% main text
\section{Introduction}

Medical image registration plays a crucial role in the field of medical image analysis by seeking to establish a meaningful connection between the voxels within two images. This process has found extensive applications in various domains, including but not limited to muscle segmentation \cite{pr_seg}, intraoperative localization \cite{mia_2021}, and quantified ablation margins and local disease progression after thermal ablation of colorectal liver metastases \cite{2023_application_registration}. By focusing on the alignment of images, medical image registration enables the exploration of changes in anatomical structures over time, providing valuable insights into patterns of variation and development.

Over the past decade, \textbf{C}onvolutional \textbf{N}eural \textbf{N}etworks (CNNs) have made significant strides in computer vision (CV), marked by their notable successes in diverse tasks \cite{Nie_1, Nie_2}. Building on this progress and the rapid evolution of CNNs, there has been a distinct shift towards CNN-based approaches in medical image analysis \cite{nnunet,2022_seg_zhengliu}. The emergence of U-Net and its variants in 2015 underscored its ability to effectively integrate both low-level and high-level semantic information while maintaining a constrained parameter count, leading to widespread adoption in various medical image analysis tasks \cite{2015-Unet,2019-unet++}. It is also employed as the backbone of the registration models, which predict the deformation fields. Driven by the long-range modeling capabilities of Vision Transformers (ViTs), recent registration researches \cite{transmorph, my,2022_xmorpher} have employed ViT blocks to leverage the modeling power. Besides using powerful modeling blocks, some methods, such as \cite{mul_reg, LapIRN}, have applied the multi-scale architecture and predicted the deformation fields at different scale stages, where the previously predicted deformation fields control the subsequent generation of the deformation fields. Unlike the multi-scale framework, the cascaded registration approaches \cite{2019-Cascade,cascaded2} have utilized a progressive manner in which a registration sub-model predicts the deformation fields based on the previous sub-model at the same resolution stage.

With the recent emergence of ChatGPT, large models have attracted widespread attention. \textbf{L}arge \textbf{L}anguage \textbf{M}odels (LLMs), such as LLaMA 2 \cite{llm_llama}, Pythia \cite{llm_pythia}, and GPT-3 \cite{llm_gpt-3} etc., which had the vast number of parameters, were trained on large-scale corpus data, aiming to understand complex linguistic content to enhance the capabilities of generating appropriate text and chatting to assist people. Recent studies \cite{vision_llm1,vision_llm2,vision_llm3} have demonstrated that pretrained LLMs are important for the cross-model (i.e., vision-language) task. The idea of the adapter appeared in the field of domain adaptation \cite{first_adapter_vision}. In order to use the powerful modeling capabilities of the pretrained model in downstream tasks, \cite{first_adapter_language} first proposed using adapters to fine-tune and apply BERT \cite{BERT} in various text classification tasks. The current LLMs are employed as sub-models in visual-language tasks to generate specific text content. Generally, to align the features from the image and language domains for downstream tasks, some methods \cite{llm_lin_pro_1,llm_lin_pro_2} utilized linear projections, i.e., adapters, to achieve the alignment of features in these two domains. 

\begin{figure}[ht!]
    \centering
    \includegraphics[width=.8\linewidth]{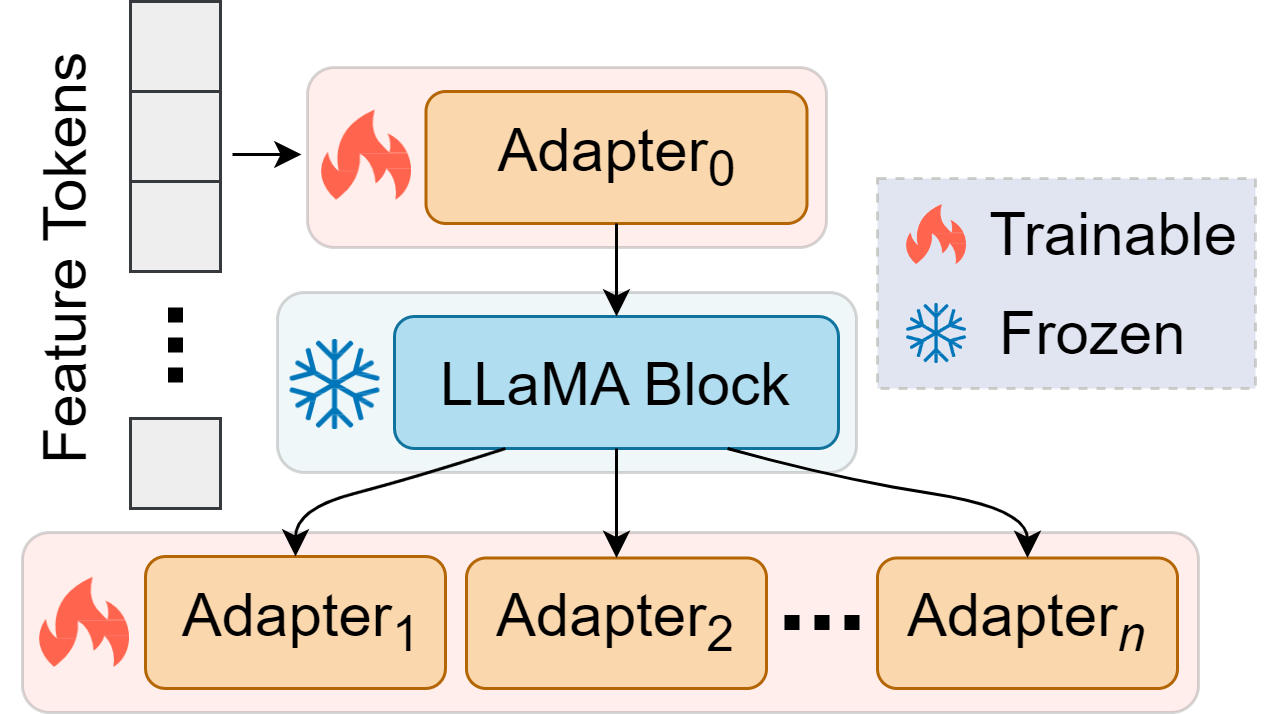}
    \caption{Overview of the utilization of LLaMA in our work. The fire icon represents the trainable block, and the snow icon indicates the frozen LLaMA block with the pretrained weights.}
    \label{fig:llama}
\end{figure}

In this paper, we propose a non-U-shaped image registration method, dubbed LLaMA-Reg, which integrates LLaMA 2 as the deep encoder for feature extraction. Our method extracts the deep features of an image pair using a two-stream CNN encoder. These features are split into feature tokens and then sent to the LLaMA encoder. The pretrained LLaMA model and its adapters are shown in Fig. \ref{fig:llama}. The LLaMA encoder of our method is frozen, which aims to boost the registration performance by using the projected features in the language domain. Before the LLaMA encoder, an adapter is utilized to project the deep features of two images to the language domain for the pretrained LLaMA encoding, as some other adapters are utilized to project the features in the language domain to the visual domain. The projected features in the visual domain are split into several branches representing different registration stages. We apply the multi-scale framework in each registration stage to achieve coarse-to-fine registration. It is worth noting that the LLaMA block in this figure is an overview. The detail of the designed LLaMA module can be seen in Sec. \ref{llama_section}.

To evaluate the performance of LLaMA-Reg, we conducted registration tasks on two anatomical image datasets: knee MR image and brain MR image. The quantitative and qualitative comparison results demonstrate the superior performance of LLaMA-Reg over state-of-the-art methods.

The summary contributions of our LLaMA-Reg are as follows:
\begin{itemize}\setlength{\itemsep}{0pt}
   \item We propose an architecture for unsupervised image registration, LLaMA-Reg, which introduces pretrained LLaMA to the registration task and employs a proposed LLaMA block to boost image registration performance.
   \item We propose a scheme for adapters in the registration task to align the language and visual domain features.
   \item We propose a non-U-shaped multi-scale and cascaded registration model to utilize LLaMA 2 for registration.
   \item The experimental results of our LLaMA-Reg on the 3D knee and brain MRI datasets demonstrate state-of-the-art performance. Ablation studies demonstrate our effectiveness in this work.
\end{itemize}

\section{Related Work}
Traditional image registration methods, such as \cite{SyN}, \cite{deeds}, and \cite{pr_otv}, used iterative optimization to find the best deformation between a pair of images. However, the problem with traditional methods is that they occupy many computing resources and are time-consuming for inference. Methods based on deep learning utilize similarity loss functions to train their weights, which can calculate the deformation field between a pair of images in a very short time after training. Unsupervised deep learning-based registration approaches have been brought to the fore since they do not require ground-truth deformation fields to train. Currently, registration methods based on deep learning that have been widely studied can be divided into two major categories: CNN-based and ViT-based.

\subsection{Convolutional Neural Networks-based Approaches}

Since the development of CNN-based registration methods, many approaches have emerged. Balakrishnan et al. first introduced VoxelMorph \cite{2019_vm}, a U-shaped structure registration model for an input pair of images that predicted full-scale displacement fields. In order to ensure the diffeomorphism characteristics of the deformation field between a pair of images and make the deformation smooth, Dalca et al. developed a diffeomorphic registration model \cite{2019-PVM}. To guarantee some other properties of deformation in registration, Mok et al. proposed SYMNet \cite{2020-SYM} to predict the bidirectional diffeomorphic deformation field. Kim et al. \cite{2021-CycleMorph} introduced the cycle consistency registration model to enhance performance and preserve topology. To further improve the performance of the registration model, Mok et al. proposed a multi-scale registration model \cite{LapIRN} based on the Laplacian pyramid network. The multi-scale model performed a coarse-to-fine registration based on the deformation field predicted by the previous scale model. \cite{NiceNet} presented a single-pass model that integrated the multi-scale scheme in the decoder to perform coarse-to-fine registration. Cascaded models, such as \cite{2019-Cascade}, employed several models to warp moving images gradually.

\subsection{Vision Transformers-based Approaches}
Since the advent of ViT, where Transformer was applied in computer vision, it has attracted the attention of many scholars in the field of medical image analysis. Chen et al. \cite{2021-vitV} integrated ViT into V-Net at the bottom of their model to perform image registration. Zhang et al. \cite{2021-zhang-DTN} introduced a dual ViT-based network to enhance the feature modeling capability. Ma et al. \cite{my} presented a symmetric variant ViT-based U-Net to improve the registration performance. Chen et al. \cite{transmorph} developed TransMorph, consisting of a Swin transformer-based encoder and a CNN-based decoder. Zhu et al. \cite{SymSwin} designed a symmetric Swin transformer-based architecture that maintains invertibility and topology preservation. In order to solve the problem that the transformation of image features to image matching relationships is implicit, TranMatch \cite{2023_transmatch}, a dual stream feature matching registration model, was proposed based on the Swin Transformer. All these ViT-based approaches mentioned above indicated performance improvement benefiting from the strong modeling power of ViTs.
% Nevertheless, the coarse-grained features computed from the transformer blocks may restrict its modeling capability. Unlike these approaches, we present an unsupervised Swin transformer-based method for deformable image registration, which enhances the contributions of the Swin transformer blocks by recovering the features and automatically builds the connections between windows.

\section{Methods}

\subsection{Image Registration}

Learning-based deformable image registration minimizes a similarity energy function to establish a dense spatial correspondence between an image pair. Given an image pair $\{I_m, I_f\}$ defined on a 3D domain $\Omega{\subset}\Bbb{R}^3$ denoting moving and fixed images. Optimization aims to find an optimal deformation field that can be formulated as
\begin{eqnarray}
        \hat{\phi}=\mathop{\arg\min}\limits_{\phi}(\mathcal{L}_{sim}({I_m}{{\circ}{\phi}},I_f)+{\lambda}\mathcal{L}_{reg}(\phi)),
            \label{eq:goal}
\end{eqnarray}
where the $I_m$ and $I_f$ denote the moving and fixed image, $I_m{\circ}{\phi}$ is the warped moving image transformed via a deformation field $\phi$. $\mathcal{{L}}_{sim}$ is the similarity matrix to estimate the similarity between $I_m{\circ}{\phi}$ and $I_f$. $\mathcal{L}_{reg}(\phi)$ is the regularization, which enforces the smoothness of the deformation field by the spatial gradient, and $\lambda$ is a hyperparameter used to balance contribution in the learning of similarity and smoothness. Hence, the optimal deformation field $\hat{\phi}$ is obtained.

In this work, we follow Eq. \ref{eq:goal} to train our deformable image registration model unsupervised. Mean squared error (MSE) is utilized as the similarity metric to evaluate the similarity between an image pair, i.e., $\mathcal{L}_{sim}= \text{MSE}({I_m}{\circ}{\phi}, I_f)$. ${\circ}$ is the spatial transform network (STN) \cite{2015-STN}, and ${I_m}{\circ}{\phi}$ represents $I_m$ warped via a deformation field $\phi$. STN can warp an image with a deformation field in an interpolation manner. We utilize the diffusion regularizer \cite{2018-VM} on the spatial gradients of a deformation field $\phi$. The regularizer is denoted as $\mathcal{L}_{reg}=\text{Diff}({\phi})$. Hence, the loss function in this work is $\mathcal{L}(I_m,I_f,{\phi})= \text{MSE}({I_m}{\circ}{\phi},{I_f})+\lambda{\text{Diff}}(\phi)$, where $\lambda$ is the hyperparameter that determines the trade-off between similarity and regularity. We optimize the parameters of LLaMA-Reg by minimizing this loss function.

\begin{figure}[ht!]
    \centering
    \includegraphics[width=1.\linewidth]{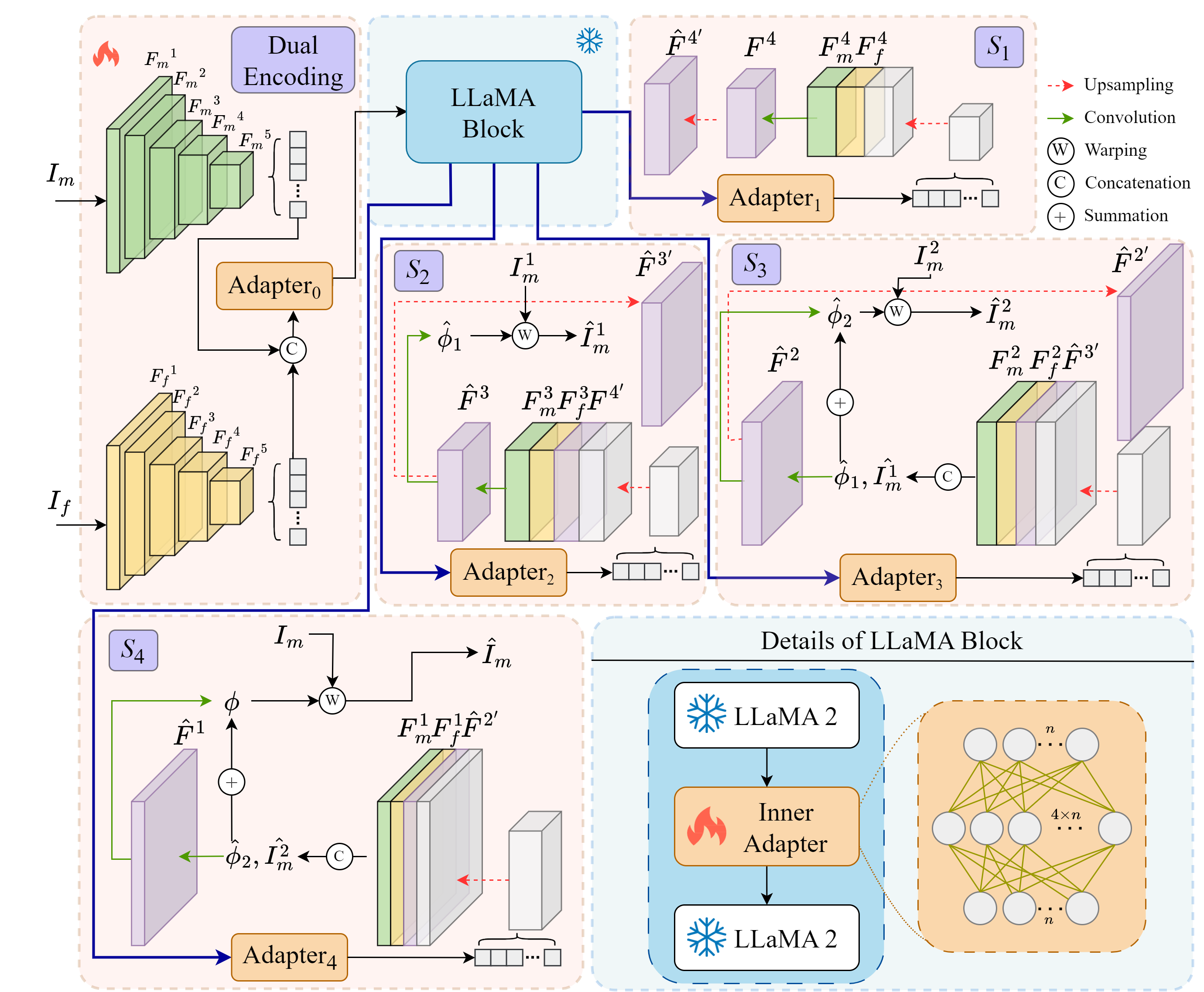}
    \caption{The overview of our proposed LLaMA-Reg. LLaMA-Reg employs a dual encoding block to extract the deep features of $I_m$ and $I_f$. Adapters are utilized to project visual features to the language domain or language features to the visual domain. $S_1$, $S_2$, $S_3$, and $S_4$ represent layers at different resolution stages from 1/8 to full resolution scale. Our model performs image registration at these stages to achieve a multi-scale registration manner. The LLaMA block is shown in the light blue box, and we offer its details in the lower right.}
    \label{fig:arch}
\end{figure}

\subsection{Adapters for Cross-Domain Projection}

For the fine-tuning methods, adapter tuning is an efficient manner to fine-tune the pretrained model with few parameters gained. In our work, we borrow the idea of adapter-tuning and use adapters composed of linear projections outside LLaMA so that the visual features obtained before LLaMA can be fitted to the parameter space (i.e., language domain) of LLaMA. More specifically, the extracted deep features of $I_m$ and $I_f$ are flattened to feature tokens. $Adapter_0$ is the first to project the $C$ dimensional features tokens in the visual domain to the language domain. It consists of two linear projections that gradually project the flattened feature tokens to the 4096 dimensions that the LLaMA block requires. The other $Adapter_i$ ($i \in {1, 2, 3, 4}$) employs single linear projection to project the encoded feature tokens in the language domain to the visual domain. Each $Adapter_i$ ($i \in {1, 2, 3, 4}$) project $C$ dimensional features tokens from LLaMA block to $2^{2(i+1)}C$ dimensions.

\subsection{LLaMA Block} \label{llama_section}

LLaMA 2 is trained on the large-scale corpus dataset, which is an open-access LLM based on the transformer. According to the number of LLaMA 2 parameters, its pretrained weights can be divided into 7B,13B, etc.
We select the 7B pretrained weight for our work. 
Methods for fine-tuning language models include adapter \cite{first_adapter_language}, LoRA\cite{hu2021lora}, etc. Adopting these technologies, the pretrained model can be applied to downstream tasks with only a small increase in the number of parameters. 
Inspired by \cite{first_adapter_vision}, we use adapters to embed LLaMA 2 into our registration model. 
Then, we design a novel LLaMA block for our registration approach.
As shown in Fig. \ref{fig:arch}, the LLaMA block consists of two pretrained LLaMA 2 that are connected by an \emph{Inner Adapter}. 
\emph{Inner Adapter} is a Multi-Layer Perceptron (MLP), a module composed of multi-layer fully connected operations that scales features in the hidden layer. 
The feature dimensions of the pretrained LLaMA 2 7B input and output are both 4096. 
The previous LLaMA 2 encodes the projected visual features, and the output features are transformed into the hidden states using the \emph{Inner Adapter}, then fed to the successive block. 
Therefore, by combining with the initial $Adapter_0$, the features in the visual domain can be converted into features in the language domain, and thus, the pretrained LLaMA2 can be applied for encoding. 
This projection and encoding can be formulated as 
\begin{eqnarray}
    {\mathcal{A}_0(x)}{\rightarrow}{x^\prime}, \,
    {{\mathcal{F}^1_{L}}(x^\prime)}{\rightarrow}{x^\prime},\,
    {\mathcal{A}_{IN}(x^\prime)}{\rightarrow}{x^\prime},\,
    {{\mathcal{F}^2_{L}}(x^\prime)}{\rightarrow}{y},
    \label{eq:proj_llama_block}
\end{eqnarray}
where $x$ is the concatenated deep features in visual domain of $I_m$ and $I_f$, $\mathcal{A}_0$ is $\text{Adapter}_0$ shown in Fig. \ref{fig:arch}, $\mathcal{A}_{IN}$ is the \emph{Inner Adapter}, and $\mathcal{F}_{L}$ is the pretrained LLaMA 2. 
According to the process of Eq. \ref{eq:proj_llama_block}, we can finally obtain feature $y$ in the language domain modeled by the LLaMA block. It is worth noting that before this block, a learnable position embedding is added to make the LLaMA block compute the visual features with its position information. 

\subsection{Registration Model with Pretrained LLaMA 2}
The proposed LLaMA-Reg and the proposed LLaMA block are shown in Fig. \ref{fig:arch}. In this work, we follow \cite{dual_prnet,NiceNet} to construct a dual-stream encoder to extract the deep features separately, which captures the semantic correspondence of two volumes independently. In the dual encoding stage, the pyramid dual encoder extracts features of $I_m$ and $I_f$. Specifically, each branch of the dual encoder has the same architecture. Both convolution blocks have a kernel size of 3. The convolution blocks with different strides of 1 or 2 are for the same resolution feature extraction or downsampling. These features are represented by ${F_m}^i \in \{{F_m}^1, {F_m}^2, ..., {F_m}^L\}$ and ${F_f}^i \in \{{F_f}^1, {F_f}^2, ..., {F_f}^L\}$, where $L$ represents the number of the resolution levels in the dual encoding stage. 

\subsubsection{Multi-Scale Registration Framework}

\begin{figure}[h]
    \centering
    \includegraphics[width=.5\linewidth]{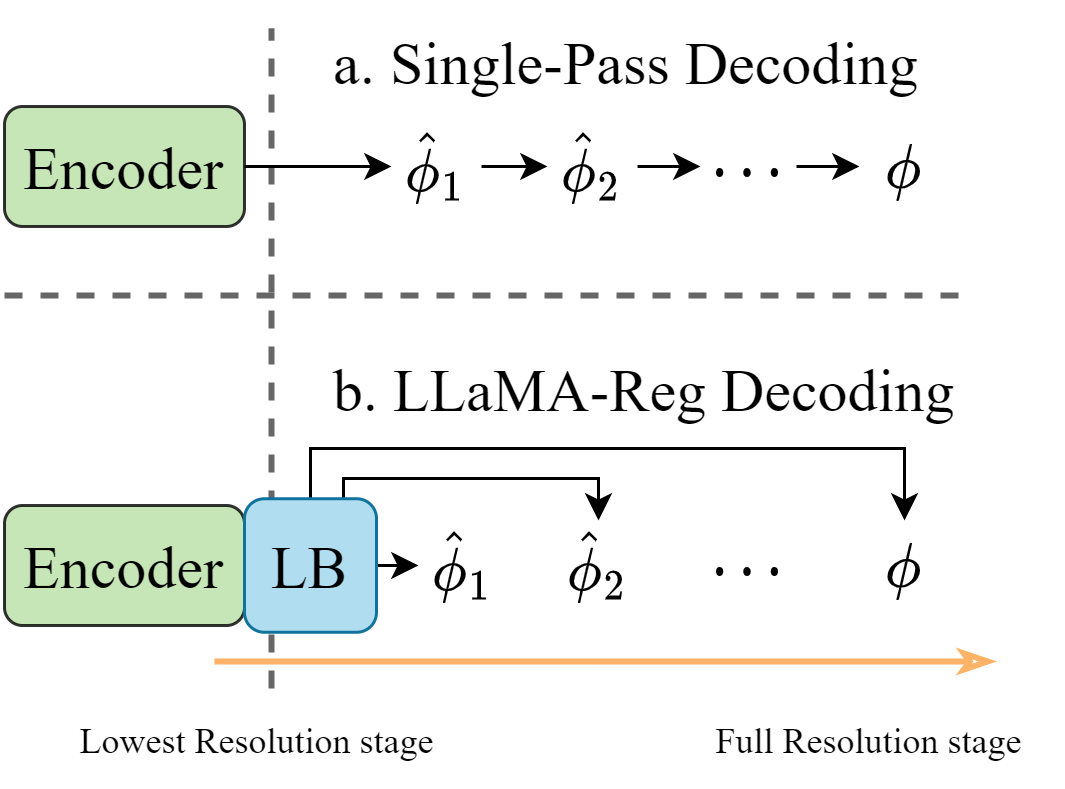}
    \caption{Two multi-scale decoding manners of Single-pass. a. is the normal decoding method, b. is the decoding method of LLaMA-Reg. LB is the LLaMA block.}
    \label{fig:diff_decoding}
\end{figure}

The multi-scale registration framework predicts the deformation field separately from low to high resolution in a coarse-to-fine manner. 
The final deformation field is predicted using the composition of deformation fields at different scales. 
The deformation field in the high-resolution stage is based on the deformation field predicted in the coarse low-resolution stage, and the more refined deformation field is predicted. 
Therefore, the multi-scale registration framework can predict a larger displacement field than the single-stage registration model \cite{LapIRN}. 
The single-pass framework \cite{NiceNet} inspired us to design a novel multi-scale registration model, LLaMA-Reg. Unlike it, as shown in Fig. \ref{fig:diff_decoding}, our registration model leverages the modeling capabilities of LLaMA to model the deep features of image pairs. At the same time, adapters are used to map features to different resolution stages and then generate deformation fields to perform multi-scale registration in the decoder.

We define $S_4, S_3, S_2$ and $S_1$ indicate the 1/8, 1/4, 1/2 and full resolution stages, respectively.
To ensure the dimensions of the output features match the number of dimensions of each resolution stage, the dimensions output by the adapter in each decoder remain consistent with the number of features of the corresponding resolution stage after reorganization.
As shown in Fig. \ref{fig:arch}, each stage utilizes the features encoded by the LLaMA block. First, stage $S_i$ obtains the encoded feature tokens in the language domain, and the $Adapter_i$ projects them into $2^{2(i+1)}C$ dimensions, where $i \in \{1, 2, 3, 4\}$.
The projected feature tokens in $S_i$ are then reconstructed to the shape of previous resolution stages, with the number of dimensions consistent with $2^{2(i+1)}C$. 
Second, in $S_i$, the reconstructed features are upsampled to the shape of the next resolution stage using transposed convolution operations with a kernel size and stride of 2. 
After upsampling, features are concatenated with the corresponding features of both $I_m$ and $I_f$ from the dual encoding stage. 
The concatenated features are computed and fused by a convolution block, followed by another convolution block to generate the deformation fields in $S_i$. 
Both convolution blocks have a kernel size of 3 and a stride of 1 to maintain the feature shape. 
Furthermore, the upsampling operation scales the fused feature to the next stage shape. 
It is worth noting that in $S_3$ and $S_4$, the deformation field and warped image obtained by composition in the previous stage are added to the feature fusion. 
In summary, based on the computation mentioned above, the calculation of the fusion of the features and previous information in each $S_i$ can be formulated as follows:
\begin{eqnarray}
   &S_1:& F^4 = \text{Conv}({F_m^4, F_f^4, F_t^4}), \hat{F}^{4^{\prime}} = \text{Up}(F^4), \\
   &S_2:& F^3 = \text{Conv}({F_m^3, F_f^3, \hat{F}^{4^{\prime}}, F_t^3}), \hat{F}^{3^{\prime}} = \text{Up}(F^3), \\
   &S_3:& F^2 = \text{Conv}({\hat{\phi}_1, \hat{I_m^1}, F_m^2, F_f^2, \hat{F}^{3^{\prime}}, F_t^2}), \hat{F}^{2^{\prime}} = \text{Up}(F^2), \\
   &S_4:& F^1 = \text{Conv}({\hat{\phi}_2, \hat{I_m^2}, F_m^1, F_f^1, \hat{F}^{2^{\prime}}, F_t^1})
   \label{eq: stage_computation} 
\end{eqnarray}
In the multi-scale registration branch, deformation fields are represented by $\hat{\phi}_i$. 
The warped moving image at resolution stage $S_i$ can be obtained by $\hat{I}_m^{i-1} = {I}_m^{i-1} \circ \hat{\phi}_{i-1}$ starting from $S_2$, where ${I}_m^{i-1}$ is the downsampled image at the corresponding stage, and $\circ$ is the warping function (i.e., STN). At $S_4$, the full resolution stage, $\phi$ is computed by a convolution block, and the final warped image is $\hat{I}_m = {I}_m \circ \hat{\phi}$. 

\subsubsection{Cascaded Registration Decoder}

\begin{figure}[h]
    \centering
    \includegraphics[width=1.\linewidth]{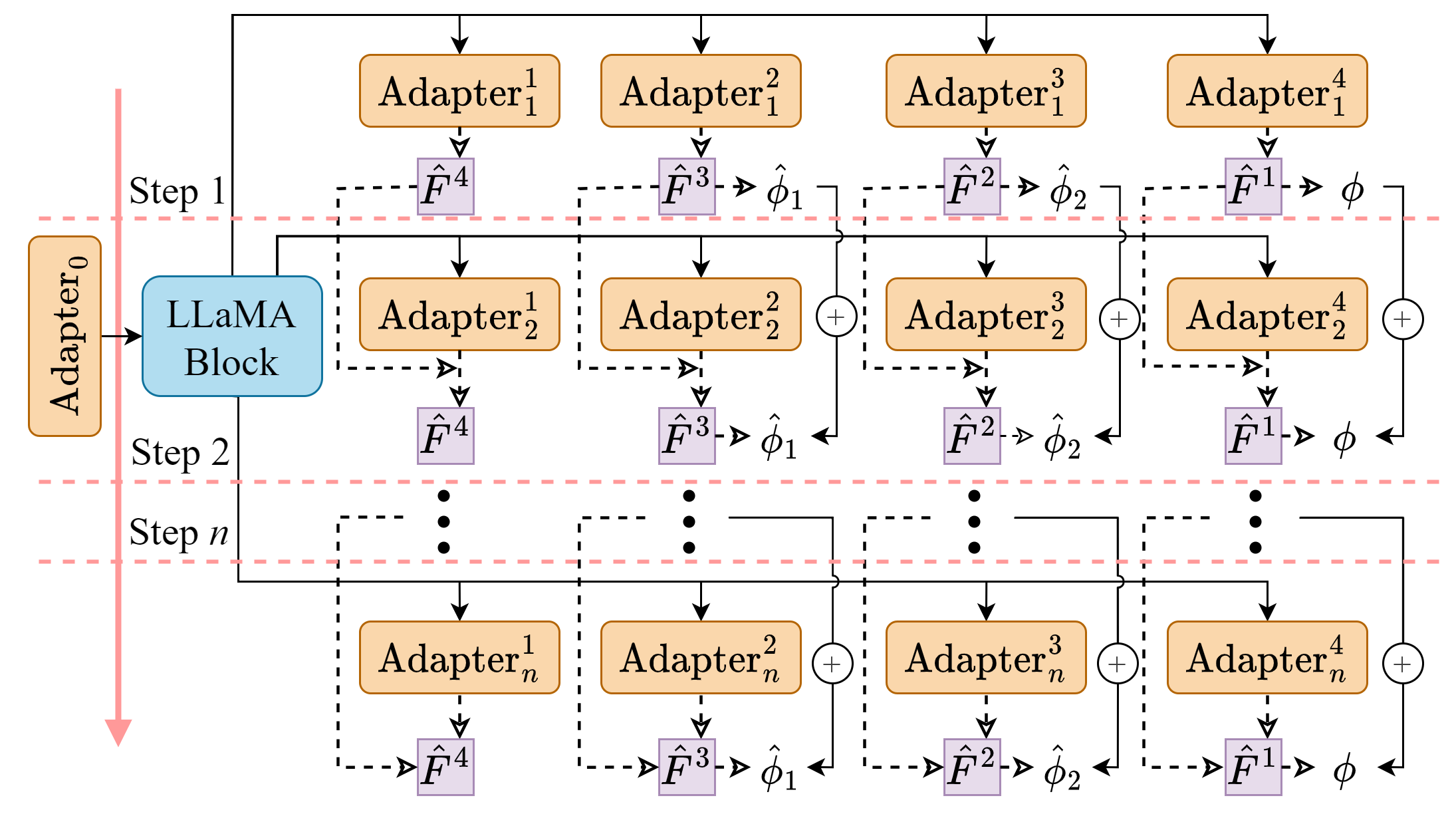}
    \caption{Overview of our proposed cascaded LLaMA-Reg. The registration procedure is split into $n$ steps. The dotted arrows indicate the omission of some intermediate calculation processes.}
    \label{fig:cas}
\end{figure}

The first model in our cascaded approach is shown in Fig. \ref{fig:arch}. 
To further enhance the registration performance, we design a novel cascaded framework. 
Different from normal cascaded image registration approaches, our cascaded LLaMA-Reg consists of the LLaMA-Reg and the cascaded decoding branches, where branches share the same dual encoding branch. 
The cascaded framework of our LLaMA-Reg is shown in Fig. \ref{fig:cas}. 
This cascaded framework splits the registration procedure into $n$ steps, and $n$ decoders indicate $n$ steps of the cascaded framework. 
These $n$ steps branches share the same pretrained LLaMA block and trained encoder in \emph{Step 1}. 
To train LLaMA-Reg efficiency and boost performance, the adapters, including \emph{inner adapters}, and decoders are unfrozen in a cascaded training phase. 
$Adapter^i_n$ is the adapter at $i$th resolution stage of step $n$. 
The feature used to generate the deformation field at each resolution stage is represented as $\hat{F}_n^{(5-i)}$. 
In the cascaded method we propose, the feature $\hat{F}_n^{(5-i)}$ generated in the previous step $n$ participates in the feature fusion process of the $n+1$ step. 
Based on this, the feature $\hat{F}_{n+1}^{(5-i)}$ generated in the next step can be obtained. 
Besides the feature fusion between features at different steps, we also added the fusion of deformation fields at different steps. 
Specifically, the deformation field predicted in the previous step directly affects the deformation field generated in the next step. 
Thus, we can obtain the final deformation field in the last step by fusing these features and superposing the deformation field.

\section{Experiments}

\subsection{Dataset Preparation}
To validate the proposed method, LLaMA-Reg, we performed subject-to-subject registration tasks using two publicly available datasets: the OAI dataset \footnote{https://nda.nih.gov/oai/} for knee MR image registration and the OASIS dataset \cite{OASIS} for brain MR image registration.

\emph{Knee Subject-to-Subject Registration.} 
For the knee subject-to-subject registration case, we employed the OAI dataset. This data repository provides images from an eleven-year longitudinal cohort study of knee osteoarthritis. Corresponding segmentation maps of knee MR images, obtained from \cite{OAI_segmentation}, were utilized to estimate registration performance. The segmentation maps include femoral bone, tibial bone, femoral cartilage, and tibial cartilage. ANTs \cite{Ants} was used to apply affine transformation to align each scan and then resample them to a size of $160\times160\times96$. The first 200 MR scans were used as the training set, and 90 MR scans were randomly selected from the remaining data (40 scans as fixed and 50 as moving images) to form a test set with a total of 2000 image pairs. For the test set, we invited a professional doctor to annotate the patella for each scan. Thus, the training set consists of 39000 image pairs, and the test set consists of 2000 image pairs, both with five labels, to train and evaluate registration performance.

\emph{Brain Subject-to-Subject Registration.}
To further demonstrate the performance of LLaMA-Reg, we conducted brain subject-to-subject registration on the OASIS \cite{OASIS} dataset, which is widely used in deep learning-based registration research. The OASIS data was obtained from Learn2Reg \cite{learn2reg}. Each image has 35 anatomical segmentation maps in OASIS. We resampled the MRI scans into the shape of $112\times160\times128$. We followed the partitioning of OASIS in Learn2Reg to train the baseline and our models. Since the test set of OASIS in Learn2Reg is not public, we used the validation set to measure the subject-to-subject registration performance. A total of 380 image pairs can be generated through the validation set.

\subsection{Baseline Methods}

We compared LLaMA-Reg with four deep learning-based methods: VoxelMorph \cite{2019_vm}, LapIRN \cite{LapIRN}, TransMorph \cite{transmorph}, and TransMatch \cite{2023_transmatch}.

VoxelMorph is a CNN-based registration model. It was the first to use a U-shaped structure to predict the full-resolution deformation field and applies the STN to warp the image through the deformation field to achieve image registration.

LapIRN is a coarse-to-fine registration model that learns multi-scale deformations. It is divided into three sub-models, predicting the deformation field from 1/4 to full resolution stages. Each sub-model is trained based on the previous model to achieve a coarse-to-fine registration process, generating large deformations.

TransMorph employs transformers to capture long-range correspondences between voxels of an image pair. It is a hybrid model consisting of CNN and Swin transformer components, demonstrating superior performance over single-scale registration methods based on CNNs.

TransMatch utilizes a dual-stream framework to encode moving and fixed images and includes a local window cross-attention module to achieve explicit feature matching. This design leverages image feature matching information to enhance inter-image matching, further improving image registration. TransMatch also uses a multi-scale registration framework.

All baseline methods were trained for 300K iterations on the knee dataset and 400K iterations on the brain dataset. Using the MSE loss function, we set the coefficient to 0.04 for the OAI dataset and 0.02 for the OASIS dataset to better fit each dataset.

\subsection{Implementation Details}
We conducted experiments on a computing platform of Ubuntu server 23 with an NVIDIA RTX 3090 GPU. Our code was implemented using PyTorch 2.0. All baseline methods were trained using the Adam optimizer, with the learning rate set to 0.0001. Our method applied the similarity loss function MSE and the spatial gradient loss regularization term to train. When the hyperparameter of the spatial regularization term was set to 0.04 and 0.02 for knee and brain image registration, LLaMA-Reg performed better. The cascaded steps of LLaMA-Reg were set to 3. LLaMA-Reg was trained in 130K, 130K, and 30K iterations on the knee dataset for each cascaded step, respectively. For the brain image registration task, each step was set to 40K, 20K, and 20K, respectively. The pretrained LLaMA 2 was obtained from the Meta website\footnote{https://ai.meta.com/resources/models-and-libraries/}, as the LLaMA-Reg deep feature encoder, which requires the number of input and output dimensions are both 4096. It is worth noting that our model and the ablations were trained using PyTorch’s automatic mixed precision strategy. Our code is publicly available at \href{https://github.com/MingR-Ma/LLaMA-Reg/}{GitHub}.

\subsection{Evaluation Metrics}

We utilized the Dice metric to measure the registration accuracy of the proposed and baseline models. The Dice metric is calculated using the overlap of the corresponding segmentation labels of two images, with values ranging from 0 to 1. A higher Dice score indicates better registration accuracy. The non-positive Jacobian determinant is used to measure folds in a deformation. If the Jacobian determinant of the deformation field is non-positive, it indicates that the area will be folded during deformation. Time consumption is calculated as the inference time required to perform registration on the GPU for each pair of images.

\subsection{Experiments}
\subsubsection{Experimental Results on OAI}

\begin{table}[htbp]
  \centering
  \caption{Comparison results of subject-to-subject knee image registration task. Higher Dice (\%) results indicate higher registration accuracy. $|J_{\phi}| \leq 0$ (\%) indicates the percentage of folding voxels in a deformation. Time represents the registration seconds consumption using a trained method. Affine Only is the initial alignment result without registration.}
\scalebox{0.8}{\begin{tabular}{cccc|cc}
    \toprule
    \multicolumn{1}{c}{\multirow{2}[4]{*}{Method}} & \multicolumn{3}{c|}{Knee}     & \multicolumn{2}{c}{Brain} \\
\cmidrule{2-6}          & \multicolumn{1}{c}{Dice} & \multicolumn{1}{c}{$|J_{\phi}| \leq 0$ } & \multicolumn{1}{c|}{Time} & \multicolumn{1}{c}{Dice} & \multicolumn{1}{c}{$|J_{\phi}| \leq 0$ } \\
    \midrule
    \multicolumn{1}{c}{Affine Only}        & 32.10      & $-$              & $-$            & 49.92 & $-$   \\
    \multicolumn{1}{c}{VoxelMorph}         & 61.63         & 0.22            & 35.19          & 76.42  &  0.17   \\
    \multicolumn{1}{c}{TransMorph}         & 64.93        & 0.29             & 65.63          & 77.53   & 0.17  \\
    \multicolumn{1}{c}{LapIRN}             & 65.26         & 0.10            & 46.79          & 77.39      &   0.10    \\
    \multicolumn{1}{c}{TransMatch}         & 66.02         & 0.19             & 82.03          &  76.67   &   0.19  \\
    \multicolumn{1}{c}{LLaMA-Reg-C1 (Ours)}& 68.98         &  0.17            & 138.81         & 77.80  & 0.18 \\
    \multicolumn{1}{c}{LLaMA-Reg (Ours)}& \textbf{71.55} &  0.26  & 241.53         & \textbf{78.40}  & 0.17 \\
    \bottomrule
    \end{tabular}}%
  \label{tab:results}%
\end{table}%

We replicated the baseline methods and retrained all methods using the same dataset partition, then compared their performance on the same test set. 
The quantitative experimental results are shown in Table \ref{tab}.
Among these methods, the registration accuracy of the early CNN-based single-scale registration model, VoxelMorph, was lower than other recent registration methods.
Among these methods, the registration accuracy of the early CNN-based single-scale registration model, VoxelMorph, was lower than that of more recent registration methods. 
Although LapIRN is also based on CNN, its accuracy was better than VoxelMorph and TransMoprh due to its multi-scale structure, which can effectively predict larger deformations. 
Using the powerful modeling capabilities of Transformers, TransMorph and TransMatch could construct distant voxel relationships, resulting in better registration performance than VoxelMorph.
TransMatch introduced a multi-layer feature matching method in the transformer, which allowed it to capture deformation information at different feature levels, thereby predicting a more accurate deformation field than other baseline methods. 
Our method, LLaMA-Reg-C1, which is the proposed model without cascaded decoders, achieved better registration accuracy on the knee dataset. 
Compared with baseline methods on the Dice metric, LLaMA-Reg-C1 outperformed TransMatch by 2.96\%, LapIRN by 3.72\%, TransMorph by 4.05\%, and VoxelMorph by 7.35\%. 
Compared with TransMatch, TransMorph, LapIRN, and VoxelMorph, the proposed LLaMA-Reg, which has three cascaded decoders, showed performance improvements of 5.53\%, 6.29\%, 6.62\%, and 9.92\%, respectively. 
By applying the cascaded decoders, which are based on multiple adapters, LLaMA-Reg showed a significant improvement, with the Dice metric increasing from 68.98\% to 71.55\% for knee scans and from 77.80\% to 78.40\% for brain scans.
\begin{figure}[htbp!]
\centering
    \includegraphics[width=0.7\linewidth]{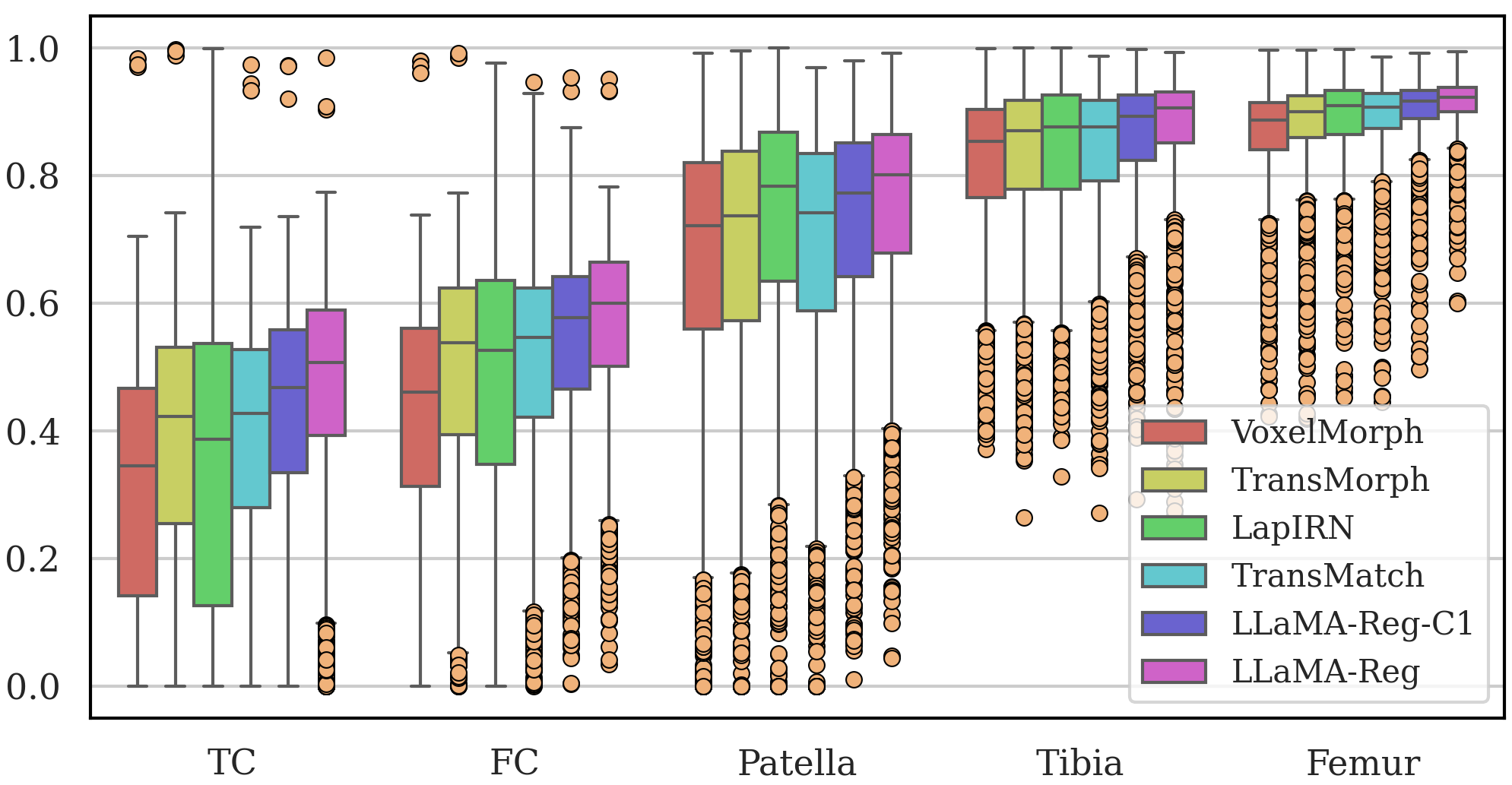}
    \caption{Box plot of registration results for all methods on each label. TC and FC represent tibia cartilage and femur cartilage, respectively.}
    \label{fig:box_oai}
\end{figure}

We presented the statistical results of all labels in Fig. \ref{fig:box_oai}. Except for the Dice results of LapIRN on the patella, which were higher than our method, ours both achieved superior registration accuracy. The experimental results on the OAI dataset showed that using the multi-scale model combined with adapters and pretrained LLaMA 2 to map the features of an image pair into the language domain could more accurately predict a deformation field. The results of $|J_{\phi}| \leq 0 (\%)$ indicated that the results of all methods are on the same order of magnitude, but the results of LapIRN were smoother. 
In the column of time consumption, we found that CNN-based methods had less inference time consumption than ViT-based methods. 
Since the more complex architectures of our methods, there was an increase in inference time.

We presented the statistical results for all labels in Fig. \ref{fig:box_oai}. Except for the Dice results of LapIRN on the patella, which were higher than our method, our methods both achieved superior registration accuracy. 
The experimental results on the OAI dataset showed that using the multi-scale model combined with adapters and the pretrained LLaMA 2 to map the features of an image pair into the language domain could more accurately predict a deformation field. 
The results of $|J_{\phi}| \leq 0 (\%)$ indicated that the results of all methods are of the same order of magnitude, but the results of LapIRN were smoother. 
In terms of time consumption, we found that CNN-based methods had less inference time than ViT-based methods. 
Due to the more complex architectures of our methods, there was an increase in inference time.

\begin{figure}[htbp!]
    \centering
    \includegraphics[width=1\linewidth]{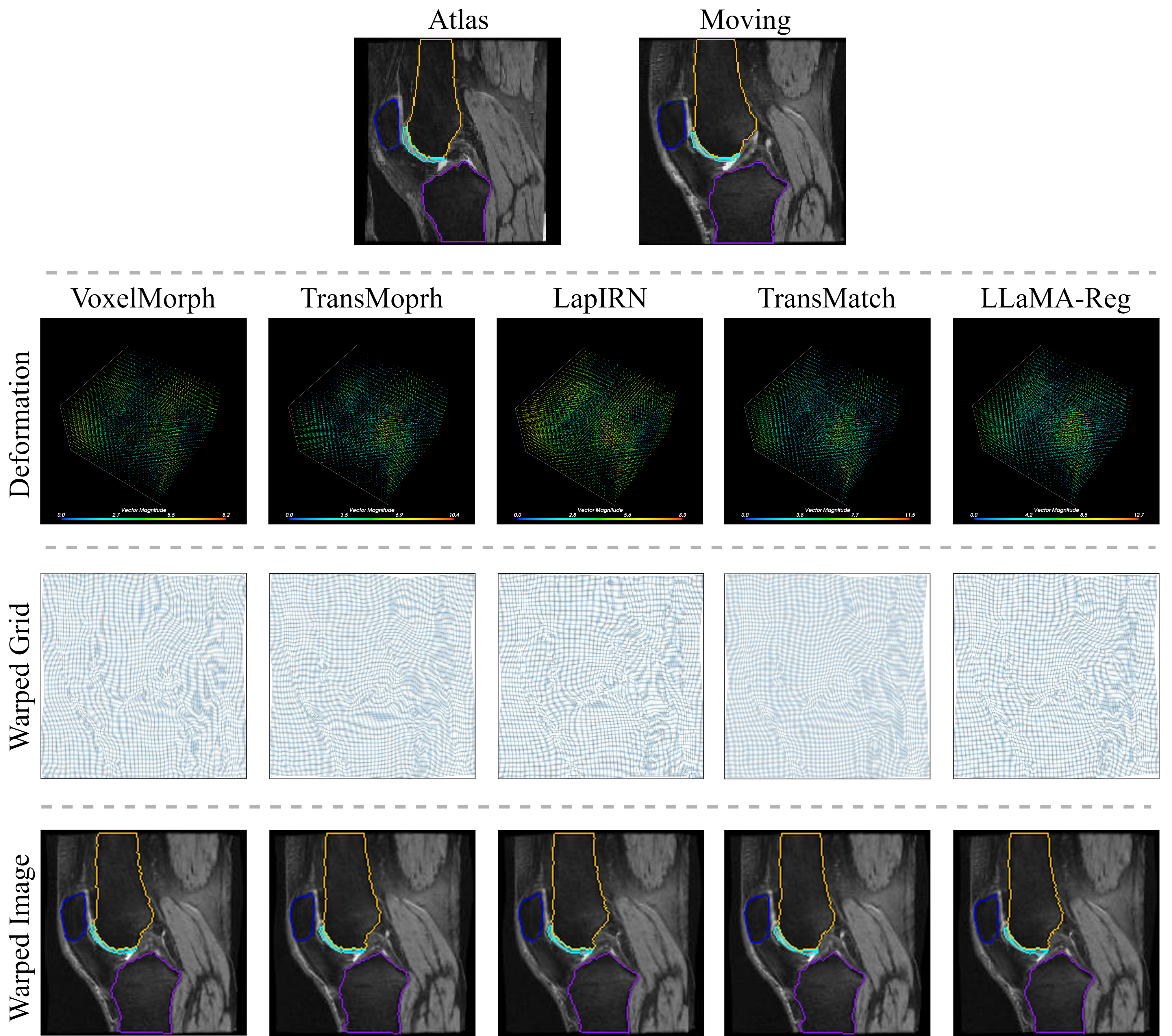}
    \caption{Visualization of experimental results on knee MRI. Deformation represents the displacement direction and magnitude of pixels in 3D MR; the warped grid reflects the changes in the current slice. In the warped images, the patella, femur, femoral cartilage, and tibia are represented in blue, yellow, light blue, and purple, respectively.}
    \label{fig:figs_show}
\end{figure}

The quantitative results are shown in Fig. \ref{fig:figs_show}. From the color bars of deformations, we noted that our method predicted the largest displacement among these methods, indicating that our method established the correspondence between more distant voxels. Through the warped images, it can be seen that among all methods, our method produced the most similar warped slice to the fixed image, especially the comparison of the bone structure represented by the labels.

\subsubsection{Additional Experimental Results on OASIS}

\begin{figure}
    \centering
    \includegraphics[width=1.0\linewidth]{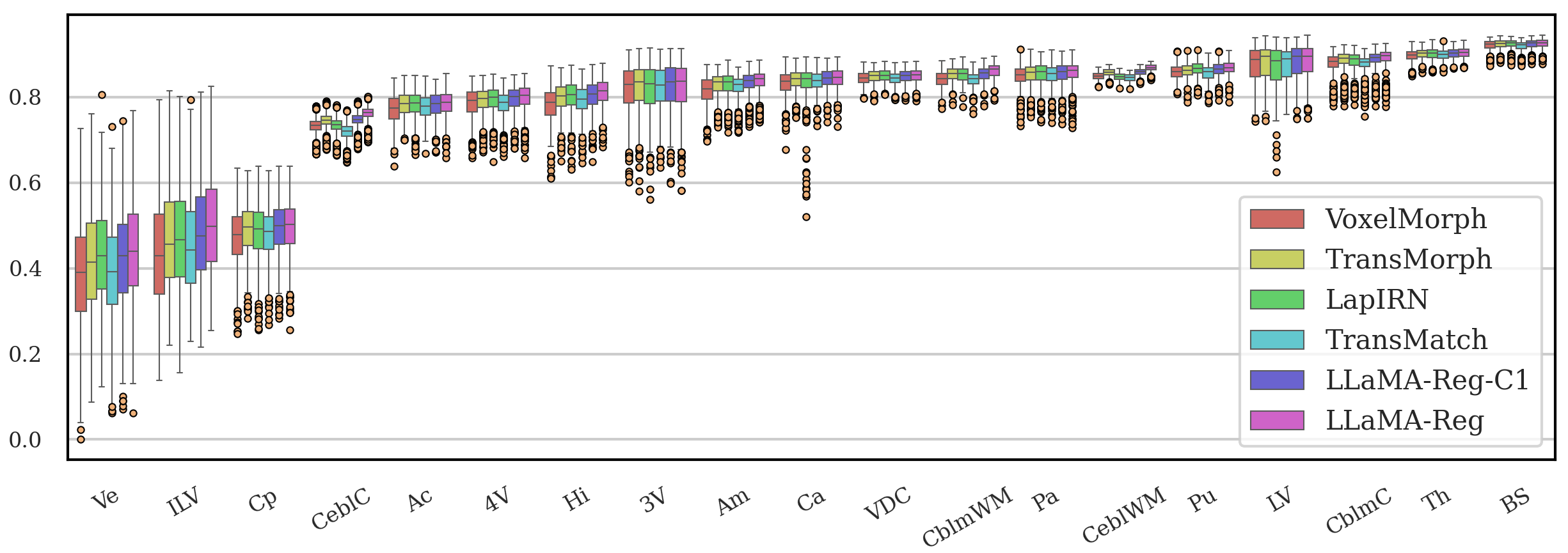}
    \caption{Box plot of registration results for all methods on OASIS.}
    \label{fig:box_oasis}
\end{figure}

We additionally tested our methods on the OASIS dataset. As shown in Table \ref{tab:results}, our methods also achieved the best registration accuracy. 
In this experiment, the result of LapIRN on the $|J_{\phi}| \leq 0 (\%)$ metric was optimal, and the results of other methods were similar. 
The result of LLaMA-Reg-C1 was better than VoxelMorph, TransMorph, LapIRN, and TransMatch by 1.38\%, 0.27\%, 0.41\%, and 1.13\%, respectively.
The registration performance of the cascaded LLaMA-Reg further improved on the brain dataset.
Compared with these methods, the performance improved by 1.98\%, 0.97\%, 1.01\%, and 1.73\%, respectively. 
By applying cascaded decoders, LLaMA-Reg had an enhancement of 0.6\%. 
We averaged the Dice scores of symmetric structures in OASIS, combining the Dice metrics of 36 labels into 19. The box plot of the statistical results on the brain is shown in Fig. \ref{fig:box_oasis}. These results illustrated that our methods performed best on most segmentation maps.

\subsection{Ablation Studies}
To verify the impact of the various proposed schemes on registration performance and their effectiveness from multiple aspects, we conducted several ablation experiments of LLaMA-Reg-C1 utilizing the OAI dataset. Ablation results are shown in \ref{tab:ablations}.

\begin{table}[htbp]
  \centering
  % \caption{Four ablation studies of our work. Ablation 1 is the experiment that compares the modeling power of the pretrained LLaMA 2 with the trainable LLaMA Transformers. The effectiveness of position embedding is in Ablation 2. Ablation 3 represents the impact of different numbers of hidden layers in the inner Adapter's MLP on the results. Results of step-by-step training and joint training for the cascaded decoder are reported in Ablation 4.}
  \caption{Three ablation studies of LLaMA-Reg-C1. Our ablation experiments report the registration accuracy and the GPU memory occupation during training.}
    \scalebox{0.8}{\begin{tabular}{c|ccccc}
    \toprule
    \multicolumn{1}{c}{} & Model & \multicolumn{1}{c}{Dice} & \multicolumn{1}{c}{Memory (MB)} \\
    \midrule
    \multirow{2}[2]{*}{Ablation 1}
          & LLaMA 2 &  67.39 &    10474     \\
          & Pretrained LLaMA 2 &  \textbf{68.98}  &   7632     \\
          & Standard ViT &  67.51 &    22520     \\
    \midrule
    \multirow{2}[2]{*}{Ablation 2} 
            & w/o Pos. Emb &   59.37    &  7208  \\
            & Dim (4096 $\times$ 4) &  \textbf{69.08}  &  7750 \\
            & Dim (4096 $\times$ 2) \& w/ Pos. Emb &   68.98  &  7632   \\
    \midrule
    \multirow{2}[2]{*}{Ablation 3} & Joint training &  71.36  &   14712   \\
          & Step-by-step &  \textbf{71.55} &   8384 (3rd Decoder)    \\
    \bottomrule
    \end{tabular}}%
  \label{tab:ablations}%
\end{table}%

\textbf{Verify the performance improvement of pretrained LLaMA 2 and model architecture.}
To demonstrate that pretrained LLaMA 2 can improve the modeling ability of the medical image registration model, we compared the registration performance of using LLaMA 2 Transformer blocks and pretrained LLaMA 2. The results of Ablation 1 indicated that the performance of the registration model using pretrained LLaMA 2 was better than using the LLaMA 2 Transformer blocks when extracting deep features. Additionally, using pretrained models can reduce memory usage and speed up training.

Furthermore, to investigate whether the scheme of our architecture had a performance advantage, we replaced LLaMA 2 blocks with standard ViT blocks. 
The configuration of ViT blocks used for replacement were: number of channels 4096, number of heads 4, number of depth 2.
The registration accuracy in Table \ref{tab:results}, showed that compared with the scheme using LLaMA Transformer blocks and pretrained LLaMA 2, the performance declined, but it still outperformed baseline methods. 
This demonstrated that our registration framework with a non-U-shaped structure could achieve higher registration accuracy, proving that the proposed non-U-shaped architecture is more suitable for registration tasks.

\textbf{Imapct of Inner settings for Performance}
When we tested the impact of the MLP mapping scale on the modeling performance of LLaMA 2, we found that although expanding the mapping scale can improve performance, the degree of improvement was limited. Therefore, we chose a mapping multiple of 2 in this work.

In the standard ViT, when using linear projection to model image features, the spatial information of these features should be modeled simultaneously. Therefore, when using transformers to solve visual problems, learnable positional encodings are added to express the spatial information of image tokens. We believe that positional embeddings need to be added when calculating image features because adapters, which consist of linear projections, need to learn spatial information. In Ablation 2, we removed the standard positional embedding in LLaMA-Reg-C1. It was observed that without positional embedding, the accuracy of registration dropped significantly. This shows that when using pretrained LLaMA 2 to calculate visual features, positional encoding is also necessary to record the positional information of the image.

\textbf{Step-by-step training and joint training for the cascaded decoder.}
We trained the cascaded decoder in two ways: step-by-step training and joint training. In step-by-step training, the previously trained network is frozen while training the next network. For joint training, the weights of the dual encoding branches were frozen, and all decoding branches were trained together. The different results of the two training manners are shown in Ablation 3. We reported the maximum GPU memory usage during step-by-step training (i.e., the 3rd cascaded step). When the number of cascaded decoding steps was set to 3, the step-by-step trained model outperformed the jointly trained model. The step-by-step trained model only needed to train the decoding part of the current step in each training stage, which occupied less GPU memory and generated more accurate deformations.

\section{Conclusion}
In this work, we propose to use the large language model LLaMA2 as the deep feature calculation component of the registration model and propose an unsupervised medical image registration model with a non-U-shaped structure.  In order to use the features calculated by LLaMA2, an adapter is used to convert visual features and language features into each other in order to transfer the calculated features to each scale stage for multi-scale registration. Experimental results on knee and brain data show that our method achieves optimal results. Through ablation experiments, we demonstrate the effectiveness of our proposed non-U-shaped registration framework and the use of pre-trained LLaMA2.

\section{Acknowledgements}
We would like to express our special gratitude to Dr. Yu Yang from the Department of Orthopedics at Taizhou Hospital for his invaluable contributions to this work, particularly in areas involving medical knowledge and medical imaging. Dr. Yang professionally annotated the test sets used for numerous experiments in the knee MRI registration task, enabling us to test all methods rigorously. Additionally, understanding the changes in bones and joints in knee imaging is crucial for analyzing and diagnosing diseases related to bones and joints. This work lays the foundation for our future research on diseases involving morphological changes in knee bones. This study was funded by the Zhejiang Province Medical Science and Technology Program of China (No. 2020PY088), the Scientific Research of Enze Medical Center (Group) (No. 24EZA01), the Scientific Research of Enze Medical Center (Group) (No. 24EZJX02), the Scientific Research of Enze Medical Center (Group) (No. 24EZCG03)
%% The Appendices part is started with the command \appendix;
%% appendix sections are then done as normal sections
%% \appendix

%% \section{}
%% \label{}

%% For citations use: 
%%       \citet{<label>} ==> Jones et al. [21]
%%       \citep{<label>} ==> [21]
%%

%% If you have bibdatabase file and want bibtex to generate the
%% bibitems, please use
%%
%%  \bibliographystyle{elsarticle-num-names} 
%%  \bibliography{<your bibdatabase>}

%% else use the following coding to input the bibitems directly in the
%% TeX file.
\bibliographystyle{elsarticle-num-names}\setlength{\bibsep}{0em}
\bibliography{egbib}
\end{document}